    \documentclass[runningheads]{llncs}
    
    \usepackage[T1]{fontenc}
    \usepackage{graphicx}
    \usepackage{amsmath}
    \usepackage{amssymb}
    \usepackage{booktabs}
    \usepackage{algorithm}
    \usepackage{algpseudocode}
    \usepackage{multirow}
    \usepackage{makecell}
    \usepackage{array}
    \usepackage[table]{xcolor}
    
    \usepackage[hidelinks]{hyperref}
\begin{document}
    
    \title{ReForest: Interpretability Meets Reasoning in Isolation Forest}
    
    \author{
    Juyoung Seo\inst{1} \and
    Geonwoo Bang\inst{1} \and
    Dongho Kim\inst{2} \and
    Moohong Min\inst{3}\thanks{Corresponding author.}
    }
        
    \authorrunning{J. Seo et al.}

    \institute{Department of Computer Education, Sungkyunkwan University, Seoul, Korea \\
    \email{\{tjwndud04, g7199\}@g.skku.edu} \and
    Department of Immersive Media Engineering, Sungkyunkwan University, Seoul, Korea \\
    \email{dpl9753@g.skku.edu} \and
    Department of Computer Education/Social Innovation Convergence Program, Sungkyunkwan University, Seoul, Korea \\
    \email{iceo@skku.edu}}
    
    \maketitle
    
    \begin{abstract}
    Statistical anomaly detection methods like Isolation Forest are efficient but suffer from a critical semantic gap. They identify numerical rarities but fail to understand the contextual meaning of data, often missing subtle, context-dependent anomalies. Conversely, Large Language Models possess strong semantic reasoning capabilities but are prohibitively expensive and slow for large-scale deployment. To bridge this gap, we propose ReForest, a novel framework that synergizes the statistical efficiency of Isolation Forest with the reasoning power of LLMs. ReForest transforms the latent decision paths of Isolation Forest into natural-language explanations, enabling an LLM to refine anomaly detection through an uncertainty-aware dynamic ensemble. This approach allows the model to adaptively weigh statistical and semantic signals based on uncertainty-driven instance-level confidence. Extensive experiments on four real-world datasets (Healthcare, Finance) demonstrate that ReForest significantly outperforms baselines, achieving an average 21\% improvement in Recall and a 10\% gain in F1-score. Our framework not only improves detection performance but also enhances model transparency by grounding semantic narratives in statistical evidence, providing interpretable rationales essential for high-stakes decision-making.
    
    \end{abstract}

    \section{Introduction}
    Anomaly detection is a cornerstone of modern data analytics, underpinning critical applications such as fraud detection in banking, intrusion detection in cybersecurity, and disease diagnosis in healthcare~\cite{ref_chandola}. Traditional statistical methods, particularly Isolation Forest (IForest)~\cite{ref_iforest}, have been widely adopted due to their linear scalability and effectiveness in identifying numerical outliers. These methods operate on the premise that anomalies are ``few and different,'' isolating them in fewer steps than normal instances.
    
    However, statistical methods face a fundamental limitation: the semantic gap~\cite{ref_skvara}. They treat features as abstract numbers, ignoring the rich semantic context and relationships inherent in real-world data. For example, in an HR dataset, a ``25-year-old employee with a \$100k salary'' might be statistically rare (an outlier), but semantically valid if their role is ``Senior Developer'' in a high-cost region. Conversely, a ``30-year-old Intern'' might be numerically normal but contextually anomalous. Purely statistical models often miss these context-dependent anomalies (False Negatives) or flag valid rare cases as anomalies (False Positives), lacking the reasoning capability to distinguish between the two.
    
    Large Language Models (LLMs)~\cite{ref_gpt4,ref_llama} have recently emerged as powerful tools for reasoning and pattern recognition, with their pre-trained world knowledge enabling contextual understanding that statistical models lack. However, applying LLMs directly to anomaly detection faces critical challenges. Processing large-scale datasets with LLMs is computationally expensive and slow, making them impractical for real-time applications~\cite{ref_xu}. Moreover, without statistical grounding, LLMs may generate plausible but incorrect justifications for anomalies, leading to unreliable predictions~\cite{ref_su}.
    
    To address these challenges, we introduce \textbf{ReForest}, a hybrid framework that combines the efficiency of Isolation Forest with the reasoning capabilities of LLMs. Unlike previous approaches that use fixed ensembles or expensive all-LLM pipelines, ReForest introduces three key innovations:
    
    \begin{enumerate}
        \item \textbf{Path-Aware LLM Integration}: We explicitly extract the \emph{isolation paths} (the sequence of rules that led IForest to flag a sample) and translate them into natural language narratives. This provides the LLM with a statistical grounding, asking it to validate the reason for isolation rather than just the raw data.
        
        \medskip
        
        \item \textbf{Uncertainty-Aware Dynamic Ensemble}: We propose a novel weighting mechanism that dynamically adjusts the influence of the statistical model and the LLM based on instance-level uncertainty. When the statistical model is confident, the system prioritizes efficiency; when it is uncertain (borderline cases), it relies more heavily on the LLM's semantic reasoning.
    
        \medskip
        
        \item \textbf{Dual-View Explainability}: Our framework provides interpretable explanations from both statistical and semantic perspectives, offering users transparent insights into why a sample is flagged as anomalous through both isolation paths and natural language reasoning.
    \end{enumerate}
    
    Extensive experiments on four real-world datasets demonstrate that ReForest effectively bridges the semantic gap. Our method successfully identifies subtle anomalies that statistical methods miss, achieving substantial improvements in Recall while maintaining comparable precision and strong F1-scores across diverse domains. Furthermore, our efficient screening mechanism significantly reduces computational costs compared to pure LLM approaches, making semantic anomaly detection practical for large-scale applications.
    
    \section{Related Work}
    \subsection{Statistical Anomaly Detection}
    Statistical and distance-based methods have long dominated the field of anomaly detection due to their computational efficiency and theoretical foundations~\cite{ref_chandola}. Isolation Forest (IForest)~\cite{ref_iforest} revolutionized the field by isolating anomalies using random trees, exploiting the principle that anomalies are easier to isolate than normal points. Local Outlier Factor (LOF)~\cite{ref_lof} measures the local density deviation of a data point with respect to its neighbors, enabling the detection of local anomalies. One-Class SVM~\cite{ref_svdd,ref_ocsvm} learns a decision boundary around normal data in a high-dimensional feature space. Distance-based methods like HBOS~\cite{ref_hbos} and KNN~\cite{ref_knn} detect anomalies based on distance metrics. More recently, ECOD~\cite{ref_ecod} uses empirical cumulative distribution functions to detect outliers without hyperparameter tuning, demonstrating strong performance across diverse datasets.
    
    Deep learning approaches have also been explored, with autoencoders, variational autoencoders, and deep SVDD~\cite{ref_deepsvdd} learning compact representations of normal data. However, these methods share a fundamental limitation with traditional statistical approaches: they lack semantic understanding. Operating purely on numerical distributions and learned embeddings, they often fail to capture complex, attribute-level relationships and domain-specific constraints that define contextual anomalies. For instance, they cannot reason about whether a high salary is appropriate given a job title and experience level.
    
    \subsection{LLMs for Anomaly Detection}
    The emergence of Large Language Models has opened new possibilities for anomaly detection through their reasoning and world knowledge capabilities~\cite{ref_gpt4}. Early works explored LLMs for time-series anomaly detection, where GPT-4 was used to interpret temporal patterns. LogGPT~\cite{ref_loggpt} applied LLMs to log analysis, leveraging their ability to understand unstructured text. In the visual domain, AnomalyGPT~\cite{ref_anomalygpt} demonstrated LLMs' capability to detect anomalies in images through vision-language models.
    
    More recently, several works have attempted to combine LLMs with traditional anomaly detection. AnomalyLLM~\cite{ref_anomalyllm} uses knowledge distillation to transfer LLM capabilities to smaller student networks, but focuses on transferring knowledge rather than real-time inference. Some approaches use LLMs as data augmentation tools or for generating synthetic anomalies~\cite{ref_yang}.
    
    However, most existing approaches suffer from one of two limitations: (1) they apply LLMs to all data points—incurring prohibitive computational costs and slow inference times for large-scale datasets, or (2) they use LLMs as black-box classifiers without leveraging the structural insights from traditional statistical models, leading to potential hallucinations and lack of statistical grounding. Furthermore, existing hybrid approaches typically use fixed combination strategies that fail to adapt to instance-level characteristics.
    
    Our work differs fundamentally by: (1) extracting and utilizing the \textit{internal decision structure} (isolation paths) from the statistical model to provide interpretable grounding for the LLM, (2) introducing an uncertainty-aware dynamic weighting mechanism that adapts the model combination strategy at the instance level rather than using fixed weights, and (3) creating a synergy where the statistical model provides efficient screening and structural insights, while the LLM validates semantic appropriateness, with each model compensating for the other's weaknesses. To our knowledge, ReForest is the first framework to systematically extract and utilize isolation path structures as interpretable features for LLM reasoning, combined with instance-level adaptive ensemble weighting based on uncertainty quantification.
    \section{Methodology}
    
\begin{figure}[ht]
        \centering
        \includegraphics[width=\textwidth]{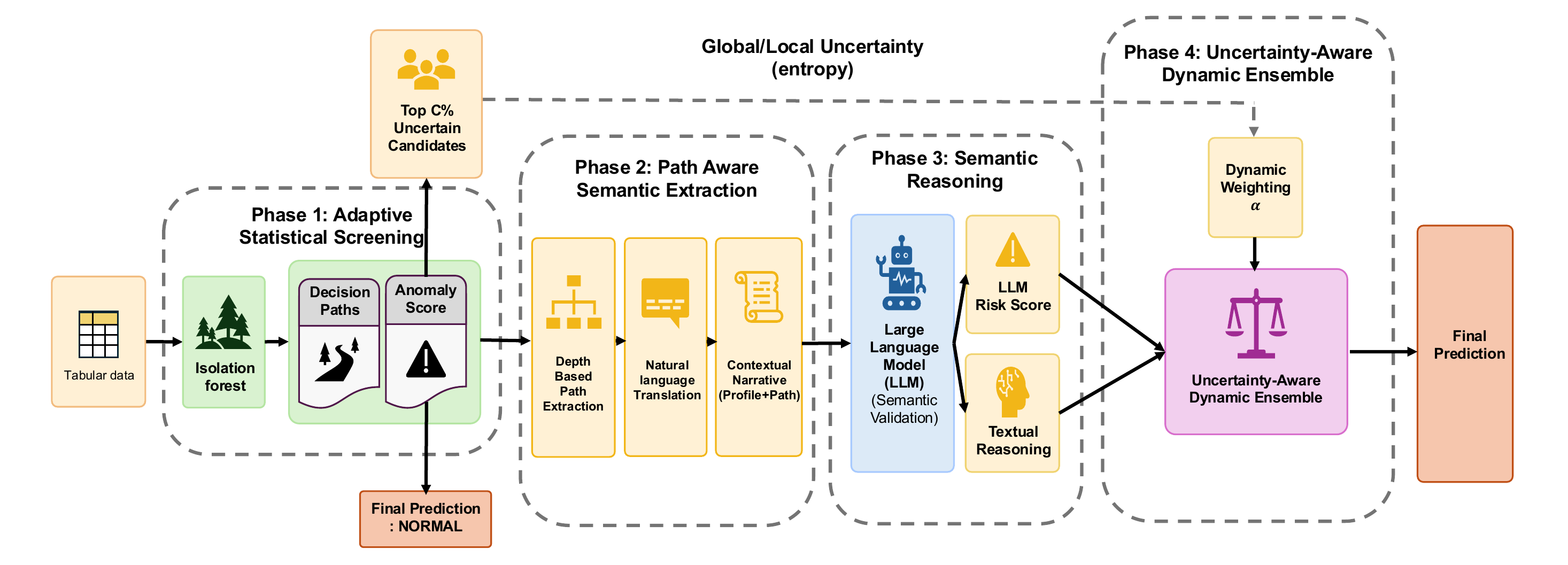}
        \caption{\textbf{The proposed hybrid anomaly detection pipeline (ReForest).} The process consists of four phases: (1) \textit{Adaptive Statistical Screening} filters out clear normal samples using Isolation Forest to reduce cost; (2) \textit{Path-Aware Semantic Extraction} selects and translates the top-$K$ shortest decision paths into natural language; (3) \textit{Semantic Reasoning} utilizes an LLM to validate anomalies based on contextual prompts; and (4) \textit{Uncertainty-Aware Dynamic Ensemble} fuses statistical and semantic scores using an \textbf{uncertainty-driven weighting mechanism $\alpha(x)$} to produce the final prediction.}
        \label{fig:reforest_overview}
    \end{figure}

    We propose \textbf{ReForest (Reasoning-Enhanced Forest)}, a hybrid anomaly detection framework that combines isolation-based structural interpretability with LLM semantic reasoning through adaptive statistical screening, diversity-aware path extraction, and uncertainty-aware hybrid scoring. The overall architectural pipeline of ReForest is illustrated in Figure \ref{fig:reforest_overview}.
    
    \subsection{Adaptive Estimation of Screening Rate}
    
    The Isolation Forest model produces an anomaly score for each sample. 
    Based on these scores, it is necessary to determine what proportion of data should be forwarded to the LLM for further inspection.
    
    Since anomaly ratios vary across datasets, we avoid using a fixed screening rate. 
    Instead, we estimate it from the training data and adjust it using a parametric safety margin and bounded constraints.
    
    Let the training dataset be:
    \[
    \mathcal{D}_{train} = \{(x_i, y_i)\}_{i=1}^{N},
    \]
    where \(y_i = 1\) indicates an anomalous sample.
    
    \medskip
    We define the estimated screening rate as:
    \[
    C_{est} = \text{Clip}\left(
    \hat{r}_{anom} \cdot \alpha,\,
    C_{min},\,
    C_{max}
    \right)
    \]
    
    \begin{itemize}
    \item \(\hat{r}_{anom} = \frac{1}{N}\sum_{i=1}^{N} \mathbb{I}(y_i = 1)\) is the estimated anomaly ratio from training data,
    \item \(\alpha\) is a safety margin hyperparameter used to cover borderline anomalies,
    \item \(C_{min}\) and \(C_{max}\) denote the minimum and maximum screening rate constraints,
    \item \(\text{Clip}(\cdot)\) restricts the final value to the range \([C_{min}, C_{max}]\).
    \end{itemize}
    
    We use \(\alpha = 1.5\) following Tukey's 1.5 IQR rule~\cite{ref_tukey}, 
    with \(C_{min} = 0.05\) and \(C_{max} = 0.50\). Samples are ranked 
    according to their anomaly scores, and only the top \(C_{est}\) 
    proportion (e.g., 100 out of 1000 samples when \(C_{est} = 0.1\)) 
    are forwarded to the LLM for further evaluation.
    
    In practice, this adaptive screening strategy provides a critical 
    computational advantage. The major bottleneck of LLM-based anomaly 
    detection is inference cost and latency~\cite{ref_su}. Using IForest as a computationally efficient pre-filter, \textit{ReForest} forwards only the top \(C_{est}\) portion of instances with the highest anomaly scores and greatest uncertainty to the LLM for semantic validation.
    
    Across our experiments, this results in an effective 
    screening ratio of approximately 15--20\%, meaning that the LLM 
    processes only a small fraction of the total samples while focusing 
    its reasoning capacity on the most ambiguous cases. This design makes 
    \textit{ReForest} practically feasible for large-scale anomaly detection 
    scenarios, avoiding the prohibitive cost of applying LLM inference to 
    every sample.

    \subsection{Depth-Based Semantic Path Extraction}
    Isolation Forest isolates anomalies through random partitioning of the 
    feature space, producing decision paths across $T$ trees. Supplying all 
    $T$ paths to an LLM would lead to information overload and excessive 
    token costs. To mitigate this, ReForest introduces a \textbf{depth-based 
    path extraction} strategy that prioritizes the most concise and 
    explanatory isolation logic.
    
    Given the set of all decision paths:
    \[
    \mathcal{P} = \{p_1, p_2, ..., p_T\},
    \]
    each path $p_t$ is represented by a sequence of split predicates, and its isolation strength is reflected by its depth $L(p_t)$. Shorter paths indicate that the instance was isolated quickly, representing a stronger and more obvious anomaly signal. Following the principle of Occam's Razor, we prioritize these shorter paths as they provide the simplest and most interpretable explanations for the LLM.
    
    We sort $\mathcal{P}$ by increasing depth and select the top-$K$ shortest paths $\mathcal{S}$ using the following procedure:
    
    \begin{algorithm}[h!]
    \caption{Depth-Based Path Extraction}
    \label{alg:depth-path}
    \begin{algorithmic}[1]
    \Require Path set $\mathcal{P} = \{p_1, \dots, p_T\}$, target number $K$
    \Ensure Selected path subset $\mathcal{S}$
    
    \State Sort $\mathcal{P}$ in ascending order of depth $L(p_t)$
    \State Initialize $\mathcal{S} \leftarrow \emptyset$
    
    \For{each $p_t \in \mathcal{P}$}
        \State $\mathcal{S} \leftarrow \mathcal{S} \cup \{p_t\}$
        
        \If{$|\mathcal{S}| = K$}
            \State \textbf{break}
        \EndIf
    \EndFor
    
    \State \Return $\mathcal{S}$
    
    \end{algorithmic}
    \end{algorithm}
    
    Each path in $\mathcal{S}$ is then translated into a natural language narrative enriched with statistical descriptions from $\mathcal{D}_{train}$ (e.g., quartile positions, extremeness). This step provides the LLM with interpretable structural signals that explain \textit{why} the statistical model flagged the instance, focusing on the most dominant isolation factors.
    
    \subsection{Uncertainty-Aware Dynamic Ensemble}
    In the final phase, ReForest fuses statistical and semantic signals through an \textbf{Uncertainty-Aware Dynamic Ensemble} mechanism. The dynamic weighting mechanism is motivated by two key insights: \textbf{(1) Statistical models excel at detecting clear-cut distributional anomalies with high confidence}, making LLM intervention unnecessary and potentially introducing noise. \textbf{(2) For borderline cases near the decision boundary, statistical uncertainty signals that distributional evidence alone is insufficient}, and semantic reasoning becomes critical. By \textbf{adaptively shifting reliance between the two modalities}, ReForest avoids both the brittleness of pure statistical methods (which ignore context) and the inconsistency of pure LLM-based detection (which may overlook statistical patterns).
    
    We normalize the IForest score using Z-score transformation and a sigmoid:
    \[
    z = \frac{\mu_{train} - S_{IF}(x)}{\sigma_{train}}, \quad P_{IF}(x) = \frac{1}{1 + e^{-z}}.
    \]
    The dynamic weight $\alpha(x)$ assigned to the statistical model is defined as:
    \[
    \alpha(x) = \text{Clip}\left(\alpha_{base} + \Delta(P_{IF}(x)),\ \alpha_{min},\ \alpha_{max}\right),
    \]
    where $\alpha_{base}$ represents the neutral weight when uncertainty is moderate, and $[\alpha_{min}, \alpha_{max}]$ defines the feasible range to prevent extreme reliance on either modality. The adjustment term $\Delta$ depends on the model's confidence:
    \[
    \Delta(p)=
    \begin{cases}
    +\beta, & |p - 0.5| > \tau_{high} \quad (\text{High confidence}) \\
    -\beta, & |p - 0.5| < \tau_{low} \quad (\text{High uncertainty}) \\
    0, & \text{otherwise},
    \end{cases}
    \]
    where $\beta$ controls the magnitude of weight adjustment, and $\tau_{high}$, $\tau_{low}$ define the confidence thresholds. In our experiments, we set $\alpha_{base} = 0.5$ (equal initial weight), $\alpha_{min} = 0.2$ and $\alpha_{max} = 0.8$ (bounded trust in either model), $\beta = 0.25$ (moderate adjustment), $\tau_{high} = 0.4$ and $\tau_{low} = 0.1$ (based on empirical calibration to capture clear vs. ambiguous regions).
    
    The final anomaly probability is:
    \[
    P_{final}(x) = \alpha(x) \cdot P_{IF}(x) + \left(1 - \alpha(x)\right) \cdot \frac{S_{LLM}(x)}{10}.
    \]
    
    The Uncertainty-Aware Dynamic Ensemble ensures that each model contributes where it has comparative advantage: statistical rigor for obvious cases, semantic judgment for ambiguous ones. This design yields three key advantages: \textbf{(1) Robustness} – by deferring to statistical confidence when available, the system avoids over-reliance on potentially inconsistent LLM outputs; \textbf{(2) Interpretability} – the extracted decision paths and LLM reasoning provide human-auditable explanations for each prediction; \textbf{(3) Context-awareness} – unlike purely statistical methods, ReForest can correctly classify rare but legitimate cases by incorporating domain semantics, significantly improving precision on context-dependent anomalies where traditional detectors generate high false positive rates.
    \section{Experiments}
    \label{sec:experiments}
    
    In this section, we validate the effectiveness of \textit{ReForest} by addressing the following three Research Questions (RQs):
    
    \begin{enumerate}
        \item \textbf{RQ1 (Performance)}: Does \textit{ReForest} outperform conventional anomaly detection models under the same experimental settings?
        \item \textbf{RQ2 (Impact of Path Information)}: How do \emph{Semantic Path Extraction} and the amount of contextual information ($K$) affect model performance?
        \item \textbf{RQ3 (Qualitative Analysis)}: Can \textit{ReForest} provide human-understandable explanations for its decisions?
    \end{enumerate}
    
    \subsection{Experimental Setup}
    \subsubsection{Datasets and Baselines}
    We utilized four datasets from diverse domains to evaluate ReForest:
    
    \begin{enumerate}
        \item \textbf{Hepatitis C Virus (HCV) Dataset} (Healthcare): Predicts diagnosis category (Blood Donor vs. Hepatitis/Fibrosis/Cirrhosis) based on clinical laboratory test results including ALT, AST, bilirubin levels, and patient demographics. Contains 615 samples with 12 features and 13.3\% anomaly ratio.
        \item \textbf{Stroke Prediction Dataset} (Healthcare): Predicts stroke risk based on medical and lifestyle factors such as hypertension, heart disease, and smoking status. Contains 4,909 samples with 10 features and 4.3\% anomaly ratio.
        \item \textbf{HMEQ (Home Equity) Loan Dataset} (Finance): Focuses on predicting default on home equity loans using features like loan amount, property value, and debt-to-income ratio. Contains 3,364 samples with 12 features and 8.9\% anomaly ratio.
        \item \textbf{Loan Default Prediction} (Finance): Predicts loan default status using borrower credit information including FICO score, interest rate, debt-to-income ratio, revolving balance, and delinquency records. Contains 9,578 samples with 13 features and 16.0\% anomaly ratio.
    \end{enumerate}
    
    We compared \textit{ReForest} against statistical baselines (IForest~\cite{ref_iforest}, LOF~\cite{ref_lof}, ECOD~\cite{ref_ecod}) and a deep learning model (AutoEncoder). All baseline implementations were based on the PyOD library~\cite{ref_pyod}.
    
    \subsubsection{Metrics}
    Given the class imbalance, we used \textbf{F1-score} as the primary evaluation metric, as it balances precision and recall. 
    In addition, we report \textbf{Recall} to emphasize the importance of detecting true anomalies, and \textbf{Precision} to assess the reliability of detected anomalies in practical applications.
    
    \subsection{Overall Performance Comparison (RQ1)}
    
    \begin{table}[h!]
    \centering
    \scriptsize
    \renewcommand{\arraystretch}{1.25}
    \setlength{\tabcolsep}{4pt}
    
    \caption{Overall performance comparison (mean $\pm$ std over available runs). Best results are shown in bold.}
    \label{tab:overall_pakdd}
    
    \begin{tabular}{l|ccc|ccc}
    \toprule
    \multirow{2}{*}{\textbf{Model}} 
    & \multicolumn{3}{c|}{\textbf{Stroke}} 
    & \multicolumn{3}{c}{\textbf{HMEQ}} \\
    & P & R & F1 & P & R & F1 \\
    \midrule
    IForest     
    & 0.18$\pm$0.01 & 0.19$\pm$0.00 & 0.18$\pm$0.00
    & 0.31$\pm$0.02 & 0.27$\pm$0.05 & 0.29$\pm$0.02 \\
    
    LOF         
    & 0.02$\pm$0.03 & 0.02$\pm$0.03 & 0.02$\pm$0.03
    & 0.15$\pm$0.01 & 0.16$\pm$0.01 & 0.15$\pm$0.00 \\
    
    ECOD        
    & 0.11$\pm$0.05 & 0.11$\pm$0.05 & 0.11$\pm$0.05
    & 0.30$\pm$0.05 & 0.27$\pm$0.02 & 0.28$\pm$0.04 \\
    
    AutoEncoder 
    & 0.09$\pm$0.03 & 0.10$\pm$0.03 & 0.09$\pm$0.03
    & \textbf{0.36$\pm$0.03} & \textbf{0.38$\pm$0.00} & \textbf{0.37$\pm$0.02} \\
    
    \midrule
    ReForest    
    & \textbf{0.18$\pm$0.01} & \textbf{0.27$\pm$0.02} & \textbf{0.22$\pm$0.01}
    & 0.24$\pm$0.00 & 0.34$\pm$0.01 & 0.28$\pm$0.00 \\
    
    \bottomrule
    \end{tabular}
    
    \vspace{0.8em}
    
    \begin{tabular}{l|ccc|ccc}
    \toprule
    \multirow{2}{*}{\textbf{Model}} 
    & \multicolumn{3}{c|}{\textbf{HCV}} 
    & \multicolumn{3}{c}{\textbf{Loan Default}} \\
    & P & R & F1 & P & R & F1 \\
    \midrule
    IForest     
    & 0.63$\pm$0.01 & \textbf{0.69$\pm$0.06} & 0.66$\pm$0.03
    & 0.28$\pm$0.00 & 0.29$\pm$0.01 & 0.29$\pm$0.00 \\
    
    LOF         
    & 0.65$\pm$0.06 & 0.63$\pm$0.10 & 0.64$\pm$0.09
    & 0.18$\pm$0.00 & 0.21$\pm$0.00 & 0.20$\pm$0.00 \\
    
    ECOD        
    & 0.65$\pm$0.03 & 0.65$\pm$0.03 & 0.65$\pm$0.03
    & 0.28$\pm$0.02 & 0.27$\pm$0.02 & 0.28$\pm$0.02 \\
    
    AutoEncoder 
    & 0.66$\pm$0.04 & 0.65$\pm$0.11 & 0.65$\pm$0.08
    & 0.28$\pm$0.02 & 0.29$\pm$0.02 & 0.29$\pm$0.02 \\
    
    \midrule
    ReForest    
    & \textbf{0.78$\pm$0.04} & 0.60$\pm$0.08 & \textbf{0.68$\pm$0.08}
    & \textbf{0.31$\pm$0.02} & \textbf{0.37$\pm$0.02} & \textbf{0.34$\pm$0.02} \\
    
    \bottomrule
    \end{tabular}
    \end{table}

    Table~\ref{tab:overall_pakdd} summarizes the overall performance of all methods across the four datasets. On the two healthcare datasets (Stroke, HCV), \textit{ReForest} achieves the highest F1-scores, improving recall on Stroke (0.27 vs. 0.19 for IForest) while maintaining comparable precision, and yielding the best F1 on HCV (0.68) through a favorable precision–recall trade-off. On the Loan Default dataset, \textit{ReForest} also attains the best precision, recall, and F1, indicating that combining statistical isolation with path-aware LLM reasoning is particularly effective when anomalies are subtle but structured. 
    
    For the HMEQ dataset, the AutoEncoder outperforms all other methods, suggesting that anomalies in this domain are characterized by complex, non-linear latent structures that are best captured by deep representation learning. \textit{ReForest} performs competitively with IForest but does not surpass the AutoEncoder, indicating that when anomalies are defined purely by numerical manifolds without substantial semantic ambiguity, the advantage of LLM-based reasoning remains limited. This distinction clarifies the specific value proposition of \textit{ReForest}: it excels in regimes that demand contextual interpretation (e.g., healthcare, loan default) rather than those dominated by purely latent numerical patterns.

    \subsection{Ablation Study: Impact of Path Information and Sensitivity Analysis (RQ2)}
    \label{sec:rq2}
    
    To jointly evaluate the contribution of semantic path extraction and 
    determine the optimal number of paths, we conducted a unified experiment 
    on the HCV dataset. We varied the number of extracted paths 
    $K \in \{1, 2, 3, 4, 5\}$ and repeated experiments across three random 
    seeds to ensure statistical reliability.
    
    \begin{figure*}[h!]
    \centering
    \includegraphics[width=\columnwidth]{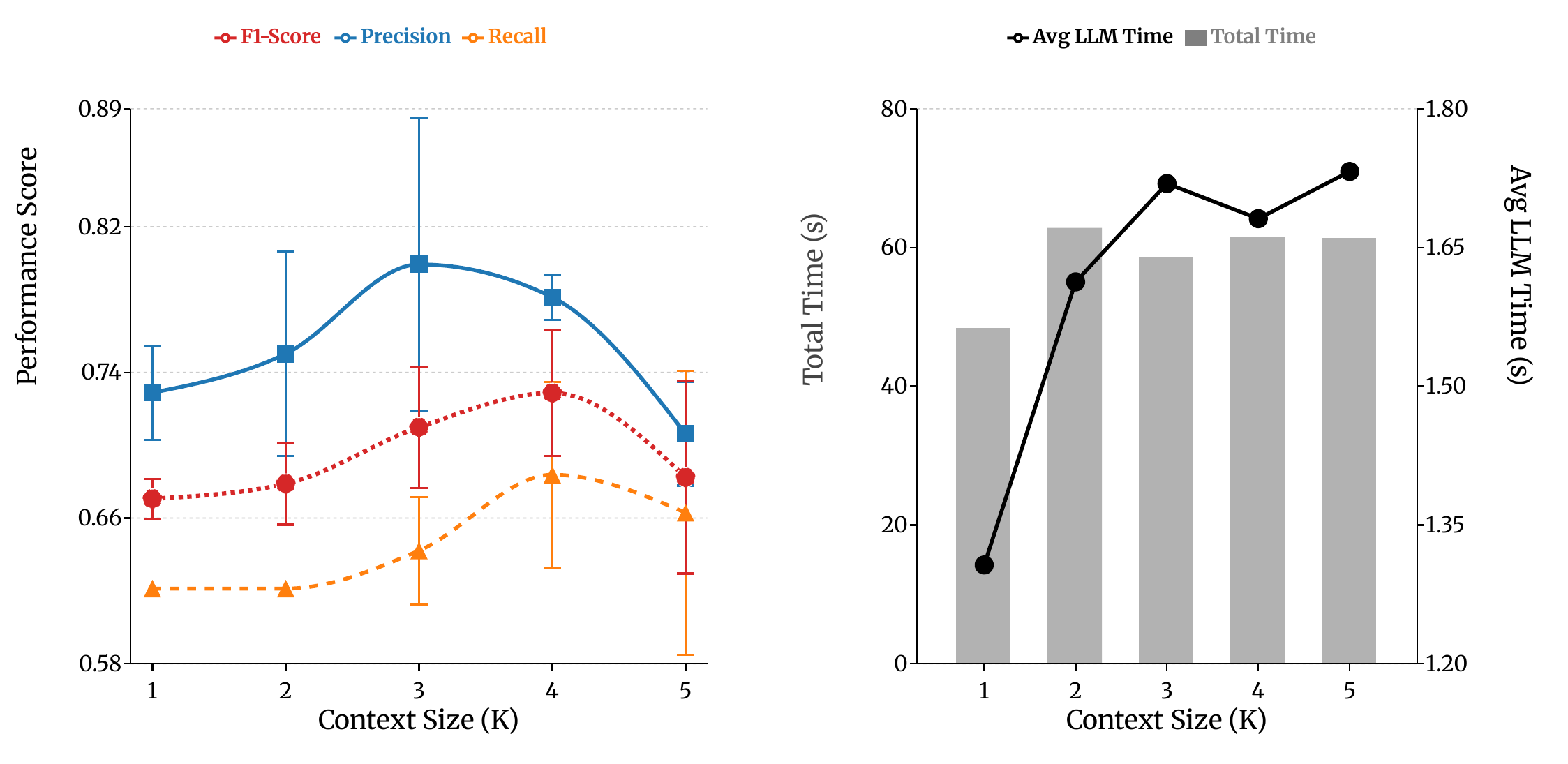}
    \caption{Top-$K$ path sensitivity analysis on the HCV dataset: performance metrics vs. $K$ (left) and inference/runtime cost vs. $K$ (right).}
    \label{fig:sensitivity}
    \end{figure*}
    
    Figure~\ref{fig:sensitivity}
    presents the sensitivity analysis results. We observe three distinct 
    regimes:
    
    \paragraph{Low Context ($K=1$--$2$):}
    For $K=1$ and $K=2$, performance gains were negligible across all three runs.
    The average F1-scores were approximately $0.674$ and $0.682$, with some runs
    showing identical values between the two settings. This indicates that a small
    number of paths fails to provide sufficient contextual diversity for improving
    the model's discrimination capability.
    
    \paragraph{Optimal Region ($K=3$--$4$):}
    Clear improvements emerged at $K=3$ and $K=4$, where the average F1-scores
    increased to approximately $0.714$ and $0.732$, respectively. Both precision
    and recall improved more consistently in this range, suggesting that moderate
    path augmentation provides useful structural context for more reliable decision-making.
    
    \paragraph{Over-Context ($K=5$):}
    When increasing to $K=5$, the average F1-score dropped to approximately $0.686$,
    and performance became less stable across runs. This degradation suggests an
    information overload effect, where excessive paths introduce noise or dilute
    attention from more discriminative signals.
    
    \subsubsection{Recommended Configuration}
    
    Based on this analysis, we adopt \textbf{$K=4$} as the default hyperparameter, 
    as it provides sufficient path diversity for robust semantic reasoning, avoids 
    overwhelming the LLM with excessive information, and minimizes inference costs 
    while maintaining optimal performance.
    
    \subsection{Qualitative Case Study (RQ3)}
    To verify the explainability and domain-specific reasoning capability of \textit{ReForest}, we analyzed representative cases where IForest misclassified samples but \textit{ReForest} correctly identified them. The cases below demonstrate how \textit{ReForest} goes beyond statistical rarity to incorporate contextual reasoning for accurate anomaly detection.

    \begin{table}[h!]
    \centering
    \caption{Qualitative Case Studies: ReForest Correcting IForest Errors}
    \label{tab:case_studies}
    \scriptsize
    \setlength{\tabcolsep}{2pt}
    \renewcommand{\arraystretch}{1.4}
    
    \begin{tabular}{@{}l l c c >{\scriptsize}l@{}}
    \toprule
    \textbf{Case} & \textbf{Sample Profile} & \textbf{Truth} & \textbf{IForest} & \textbf{ReForest Reasoning} \\
    \midrule
    
    \makecell[l]{Case 1:\\HCV\\(FP Correction)} & 
    \makecell[l]{\texttt{Age=50}, \texttt{Sex=M}\\\texttt{BIL=4.5} (Low)\\\texttt{CHOL=8.8} (High)\\\texttt{GGT=239} (High)} &
    Normal &
    \makecell{Anomaly\\(1.54)} &
    \makecell[l]{\textbf{Normal (2/10)}\\``Elevated CHOL/GGT are isolated\\ deviations. No signs of acute liver failure''} \\
    \midrule
    
    \makecell[l]{Case 2:\\Stroke\\(FP Correction)} &
    \makecell[l]{\texttt{Age=73}, \texttt{BMI=37.7}\\\texttt{Glucose=230.7}\\\texttt{HeartDis=1}\\\texttt{Hyperten=1}} &
    Normal &
    \makecell{Anomaly\\(1.68)} &
    \makecell[l]{\textbf{Normal (2/10)}\\``High risk factors (Glucose, BMI) present,\\but likely managed. No acute stroke event.''} \\
    \midrule
    
    \makecell[l]{Case 3:\\HMEQ\\(FP Correction)} &
    \makecell[l]{\texttt{LOAN=\$40.3k}\\\texttt{VALUE=\$467k}\\\texttt{MORTDUE=\$370k}\\\texttt{DELINQ=0}} &
    Normal &
    \makecell{Anomaly\\(1.62)} &
    \makecell[l]{\textbf{Normal (3/10)}\\``High loan amount is well-collateralized by\\high property value. Clean credit history.''} \\
    \midrule
    
    \makecell[l]{Case 4:\\HMEQ\\(FN Detection)} &
    \makecell[l]{\texttt{LOAN=\$18.2k}\\\texttt{DEBTINC=78.7\%}\\\texttt{DELINQ=3}\\\texttt{DEROG=0}} &
    Anomaly &
    \makecell{Normal\\(1.51)} &
    \makecell[l]{\textbf{Anomaly (7/10)}\\``Critical risk: Extreme Debt-to-Income\\(78\%) combined with delinquency history.''} \\
    
    \bottomrule
    \end{tabular}
    \end{table}

    These cases demonstrate that \textit{ReForest} goes beyond simple statistical rarity by integrating \textbf{domain knowledge} and \textbf{causal reasoning} to correct misclassifications. Cases 1--2 illustrate False Positive Correction, where statistically rare patterns are identified as legitimate business contexts. Cases 3--4 showcase False Negative Detection, where individually ordinary variables form contextually suspicious combinations. Consequently, \textit{ReForest} not only improves detection performance but also provides human-comprehensible \textbf{rationales}, significantly enhancing the trustworthiness of the AI system.

    \section{Conclusion}
    We presented \textit{ReForest}, a hybrid anomaly detection framework that couples the statistical efficiency of Isolation Forest with the semantic reasoning capability of large language models. By extracting isolation paths as interpretable structural signals and combining them with an uncertainty-aware dynamic ensemble, ReForest adaptively balances numerical anomaly scores and contextual validation.
    
    Experiments on four real-world healthcare and finance datasets show that ReForest improves recall and F1-score in settings where anomalies are context-dependent, while an adaptive screening strategy keeps LLM usage to roughly 15–20\% of samples, making the approach computationally practical. At the same time, our analysis reveals that the benefit of semantic path information is dataset-dependent, and can be limited when anomalies are primarily driven by clear statistical patterns.
    
    Future work includes refining uncertainty calibration to better decide when semantic reasoning is necessary, augmenting prompts with structured domain knowledge, and extending ReForest to multimodal and streaming scenarios with more cost-aware semantic routing. Overall, ReForest moves anomaly detection beyond purely numerical rarity toward more interpretable, context-aware decision-making suitable for high-stakes real-world deployment.
    
    \section*{Acknowledgements}
    This research was supported by the Institute of Information \& Communications Technology Planning \& Evaluation (IITP) grant funded by the Korea government (MSIT) (No. RS-2024-00439139, Development of a Cyber Crisis Response and Resilience Test Evaluation Systems). This research was also supported by the MSIT (Ministry of Science and ICT), Korea, under the Graduate School of Virtual Convergence support program (IITP-2026-RS-2023-00254129) supervised by the IITP (Institute of Information \& Communications Technology Planning \& Evaluation).
    
    \bibliographystyle{splncs04}
    
    \end{document}